\documentclass[10pt]{iopart}

\usepackage{graphicx}
\usepackage{multirow}

\begin{document}

\title[OpenMPR]{OpenMPR: Recognize Places Using Multimodal Data for People with Visual Impairments}

\author{Ruiqi Cheng \& Kaiwei Wang \& Jian Bai}
\address{State Key Laboratory of Modern Optical Instrumentation, Zhejiang University, Hangzhou, China}
\ead{wangkaiwei@zju.edu.cn}
\author{Zhijie Xu}
\address{School of Computing and Engineering, University of Huddersfield, Queensgate, Huddersfield, UK}
\vspace{10pt}
\begin{indented}
\item[]November 2018
\end{indented}

\begin{abstract}
Place recognition plays a crucial role in navigational assistance, and is also a challenging issue of assistive technology. The place recognition is prone to erroneous localization owing to various changes between database and query images. Aiming at the wearable assistive device for visually impaired people, we propose an open-sourced place recognition algorithm OpenMPR, which utilizes the multimodal data to address the challenging issues of place recognition. Compared with conventional place recognition, the proposed OpenMPR not only leverages multiple effective descriptors, but also assigns different weights to those descriptors in image matching. Incorporating GNSS data into the algorithm, the cone-based sequence searching is used for robust place recognition. The experiments illustrate that the proposed algorithm manages to solve the place recognition issue in the real-world scenarios and surpass the state-of-the-art algorithms in terms of assistive navigation performance. On the real-world testing dataset, the online OpenMPR achieves 88.7\% precision at 100\% recall without illumination changes, and achieves 57.8\% precision at 99.3\% recall with illumination changes. The OpenMPR is available at https://github.com/chengricky/OpenMultiPR. 
\end{abstract}

%
\vspace{2pc}

%
%
\ioptwocol

\section{Introduction}
\label{intro}
Vision provides people with the majority of environmental information. Up to 253 million people in the world are with visual impairments~\cite{WHO2017}, and they encounter various difficulties in their daily life. The visually impaired people have limited capability to acquire spatial knowledge~\cite{SpatialNavigation}, hence visual place recognition is desired by the visually impaired people, especially in the complex and unfamiliar outdoor environments.

Among the decades, GNSS (global navigation satellite system) has become a prevailing approach to positioning in many applications, such as vehicle navigation, engineering measurement and etc. In order to promote the positioning performance, a number of GNSS processing methods~\cite{GNSS1,GNSS2,GNSS3,GNSS4,GNSS5} were proposed by the research community to reduce the localization error up to even several millimeters. However, on the low-cost portable devices, the performance of GNSS localization is usually insufficient for the localization demands of the visually impaired people. Compared with that, optical images containing extra positioning cues could be exploited to achieve precise localization. Leveraging images to localize is known as \textit{place recognition}, which is to select the corresponding image of a given query image from database images.

The challenging issues of place recognition lie in applying the place recognition algorithm to real-world scenarios, where the visual appearance of query and database images suffers from variations, such as illuminance changes and viewpoint changes~\cite{VisualLocalizer}. With the proliferation of computer vision, the challenging place recognition task has attracted many researchers to make contributions in this area. 
Apart from the appearance changes between database and query images, the navigational assistance for people with visual impairments brings in more challenges for the task of place recognition. In the research area of intelligent vehicles, the stationary car-mounted cameras capture the images with high resolution and large field of view, and the accuracy of around several tens of meters is sufficient for car localization. However, the images captured by the wearable devices usually feature low quality, such as the severe motion blur and the continuously changing viewpoint. Moreover, assistive navigation requires more accurate localization, especially at some key positions like street corners, gates and bus stations. 

In our previous work~\cite{KeyPosition}, multimodal images and GNSS data were used to achieve key position prediction, which aimed to localize the visually impaired person at the positions of interest. Besides, we also implemented Visual Localizer~\cite{VisualLocalizer}, which utilized CNN (convolutional neural network) descriptor and data association graph to achieve place recognition for visually impaired people. Aiming at the scenarios of assistive technology, we propose a real-time place recognition algorithm OpenMPR (open-source multimodal place recognition), which extends our preceding research. In this paper, the multiple descriptors of multimodal data and parameter tuning schemes are incorporated to robustify the performance of place recognition in real world. Compared with existing algorithms, OpenMPR runs in an online fashion that only the ``past'' query images are utilized for place recognition, hence it could be used on wearable assistive devices in real time.

\begin{figure} 
	\begin{minipage}{\columnwidth}
		\centering
		\includegraphics[width=\textwidth]{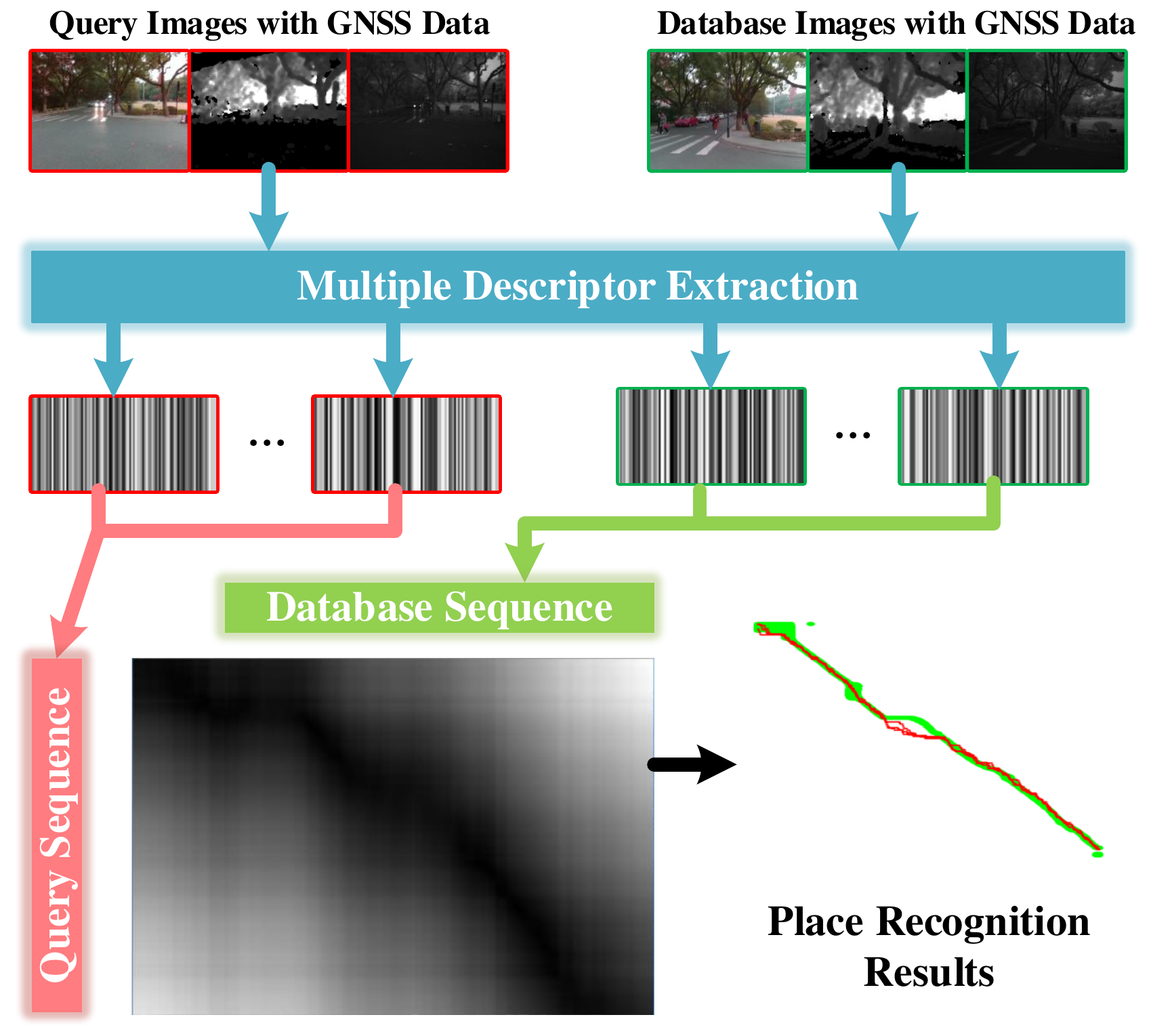}
	\end{minipage}
	\caption{The schematic diagram of OpenMPR, an open-source multimodal place recognition algorithm proposed in this paper.}
	\label{fig:1}
\end{figure}

The place recognition procedures of OpenMPR are shown in Figure~\ref{fig:1}. Multiple descriptors are extracted from multimodal data in both database and query sequences, and the multiple distance matrices are subsequently calculated. Subsequently, the score matrix is synthesized by the distance matrices of different modal data. Finally, the place recognition results are selected from the candidates with high matching scores. The contributions of this paper are summarized as follows:
\begin{itemize}
	\item  To cope with the appearance changes in place recognition, multimodal data, including the images of different modalities and GNSS data, are leveraged for place recognition tasks.
	\item  In order to exploit the latent ``place fingerprint'' embedded in those data, training-free multiple image descriptors are utilized. The weights of those descriptors are tuned to improve the performance of place recognition.
	\item  Aiming at tackling the localization issues of people with impaired vision, we propose an online place recognition algorithm OpenMPR that surpasses the state of the art and a place recognition dataset collected in the real-world scenarios.
\end{itemize}

This paper is organized as follows. The related work on place recognition is described briefly in Section~\ref{RelatedWork}. The place recognition algorithm based on multimodal data implemented in OpenMPR is presented in Section~\ref{OpenMPR}. Moreover, the comprehensive performance experiments are detailed in the Section~\ref{Experiments}. The last section concludes the paper and presents future work.

\section{State of the Art}
\label{RelatedWork}
Place recognition is a prevalent research topic among the communities of computer vision and robotics. According to the types of map abstraction, visual localization falls into metric place recognition and topological place recognition~\cite{VPR_SURVEY}. Metric place recognition returns localization results with metric information. It includes various SLAM (simultaneous localization and mapping) systems (e.g. ORB-SLAM2~\cite{ORB-SLAM2}) and deep pose prediction networks (e.g. PoseNet~\cite{PoseNet}). Although SLAM systems build the three-dimensional metric maps which could be reused to estimate precise camera poses, they are not suitable for visual localization in changing and large-scale outdoor environments. The deep networks though feature superior robustness against appearance changes, they need to be trained exclusively for each region to predict camera poses in that specific region. For building metric maps, video streams are required as input data to ensure enough scene overlap between successive frames, which is not necessarily available to the wearable assistive devices with limited computational resources. Therefore, metric place recognition is not the optimal choice for assistive technology. Avoiding to build metric maps, topological place recognition generates localization results without metric information. Topological place recognition is suitable for assistive navigation, considering it does not require high-performance hardware and ideal environments.


The community of autonomous vehicles has developed numbers of algorithms to pursue better performance on topological place recognition. Using bag-of-word method, OpenFAB-MAP~\cite{OpenFABMAP} is one of earliest open-source packages to achieve appearance-based place recognition. Different kinds of data were leveraged in the existing place recognition algorithms. GNSS priors were exploited in the computationally expensive matching process based on minimum network flow model~\cite{GPSPrior}. Sequence-based LDB (local difference binary) features derived from intensity, gradient and disparity images were utilized to depict images and achieved life-long visual localization in OpenABLE~\cite{OpenABLE}. However, the multimodal LDB descriptors were simply concatenated into single image feature, thus the weights of different modalities in place recognition were not considered. Multiple descriptors were leveraged to achieve sequence-based image matching~\cite{Han2018}, but only color images were used as visual knowledge. Taking advantage of sequence search and match selection, Open-SeqSLAM2.0~\cite{OpenSeqSLAM2.0} designed configurable parameters to explore the optimal performance of place recognition under changing conditions. 

The appearance variations impede the performance of visual place recognition, and many researchers are dedicated to mitigating the impact of appearance variations towards place recognition by different methods~\cite{VisualLocalizer,KeyPosition,PoseNet,netVLAD}. The illumination change is one of vital appearance variations, and quite a few place recognition algorithms~\cite{IlluminationInvariantImaging,ChangeRemoval} addressed the issue. Illumination invariant transformation was proposed to improve visual localization performance during daylight hours~\cite{IlluminationInvariantImaging}. Change removal based on unsupervised learning was utilized to achieve robust place recognition under day-to-night circumstances~\cite{ChangeRemoval}. Despite the fact that inspiring progress has been obtained by those work, there are challenging issues to be addressed on place recognition for assistive navigation, which has not aroused the sufficient attention of the research community.

To evaluate the performance of place recognition, substantial datasets were proposed by the research community, and some typical datasets feature different appearance variations between query and database images. Those datasets involve the cross-season Nordland dataset~\cite{Nordland} as well as Gardens Point Walking dataset~\cite{Garden} with viewpoint and illuminance variations. Bonn dataset~\cite{Bonn} and Freiburg dataset~\cite{GPSPrior} both feature multiple variations, including season, illuminance and viewpoint changes. Most of the datasets are designed for place recognition on autonomous vehicles, the images captured by car-mounted cameras are different with those captured by wearable devices. Besides, the ground truths of those datasets are labeled with GNSS data, hence the localization resolution is not sufficient for assistive technology. To the best of our knowledge, the dataset with multimodal images for assistive technology has not been released. 

\section{OpenMPR}
\label{OpenMPR}
Different from the existing place recognition approaches, OpenMPR leverages multimodal data to address the issues of place recognition. Apart from vanilla color images, other visual modalities (i.e. depth images and infrared images), as well as GNSS data, are also considered in the system. Multiple descriptors are utilized to exploit the latent information embedded in the multimodal images. The distance matrices derived from the multiple descriptors of query and database images are merged into a synthetic score matrix. Subsequently, the sequence-based matching and selection are executed to obtain the final place recognition results.

\subsection{Multiple descriptors extraction from multimodal images}
\begin{figure*}
	\centering
	\includegraphics[width=\textwidth]{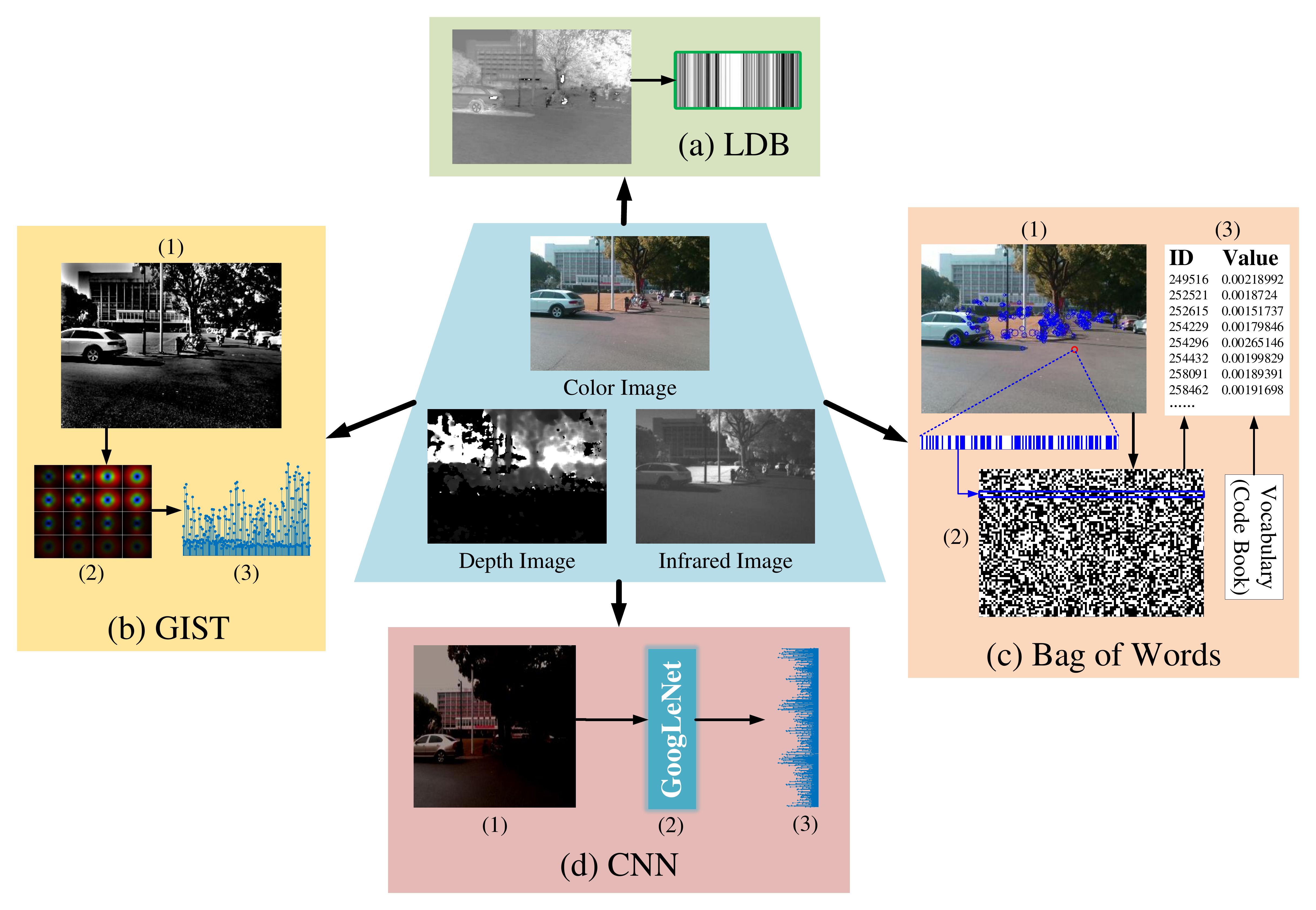}
	\caption{The multiple descriptors extracted from the multimodal images.}
	\label{fig:2}
\end{figure*}

The multimodal images involved in OpenMPR are color images, depth images and near-infrared images. The vanilla color image is an indispensable modality in place recognition task, in that it conveys the both holistic scenes and local textures with chromatic visual cues. Compared with color images, infrared images occupy a longer-wavelength band in spectrum, thus naturally carry different scene information. Depth images contain the three-dimensional shapes, which reduce the odds of mismatching between query and database images. In order to describe the scenes comprehensively, not only are the multimodal images captured to enrich the input information but also both holistic and local image descriptors are utilized to extract the key visual cues embedded in images. As shown in Figure~\ref{fig:2}, four training-free and pre-trained descriptors are chosen to depict scenes, which avoids the training procedures toward the regions to be deployed so as to be applied to assistive navigation. 

The descriptor vector extracted by descriptor $f$ from modality $m$ is defined as $ \textbf{d}^{f,m} $ in this paper. The descriptor $f$ could be one of descriptors in the set
\begin{equation}
F=\{f|GIST,~LDB,~BoW,~CNN\},
\end{equation}		
and the modality $m$ could be one of modalities in the set
\begin{equation}
M=\{m|color,~depth,~infrared\}.
\end{equation}
The concrete extraction configurations of those descriptors have been illustrated in our previous work~\cite{KeyPosition,VisualLocalizer}. Herein, we summarize descriptor extraction as follows.

\subsubsection{Bag of words}
Based on the local feature ORB (oriented FAST and rBRIEF)~\cite{ORB}, BoW (bag of words) characterizes the image details by the occurrence of each visual word clustered by local features. BoW is widely applied to object and scene categorization, due to its simplicity, computational efficiency and invariance to affine transformation~\cite{BoW}. In this paper, the key points [see Figure~\ref{fig:2} (c1)] are detected by oriented FAST (features from accelerated segment test) and are described by rBRIEF (rotated binary robust independent elementary features). The ORB descriptors of all key points are merged together and compose the concatenated descriptors [see Figure~\ref{fig:2} (c2)]. Subsequently, the BoW descriptor [see Figure~\ref{fig:2} (c3)] is generated using the extracted ORB descriptors and the pre-trained vocabularies~\cite{DBoW3}. In view that the off-the-shelf vocabularies were trained on photometric images, BoW desceiptors are extracted from color and infrared modalities.

\subsubsection{Local difference binary}
The holistic image descriptors, i.e. GIST~\cite{GISTOliva2001,GIST1238354}, LDB~\cite{LDB} and CNN descriptors, emphasize whole visual features rather than local details, hence are used to alleviate the impact of appearance changes for image matching. As shown in Figure~\ref{fig:2} (a), LDB descriptor is extracted as a global descriptor after the preprocessing of illumination invariance transformation. It is worthwhile to note that bit selection~\cite{LDB} is not executed in this paper, in that the compression of the global descriptor hinders the performance of image description. LDB descriptors are extracted from all of the modalities separately.

\subsubsection{GIST}
Also as a holistic image descriptor, GIST represents the scene in a very low dimensions. The global descriptor GIST is extracted from the preprocessed image, which involves image normalization [see Figure~\ref{fig:2} (b1)], Gabor filtering [see Figure~\ref{fig:2} (b2)] and response averaging [see Figure~\ref{fig:2} (b3)]. GIST descriptors are extracted from all of the modalities separately.

\subsubsection{CNN}
Different from the hand-crafted descriptors above, the descriptors selected from CNN are also used to enhance the description ability of the system. As presented in Figure~\ref{fig:2} (d), the CNN descriptor is generated from the intermediate layers of the pre-trained GoogLeNet fed by the preprocessed color image. The compressed concatenation of two layers $ inception3a/3\times3 $ and $ inception3a/3\times3\_reduce $ in GoogLeNet pre-trained on Places365 dataset~\cite{places} is used for image descriptor. Constrained by the structure of GoogLeNet, the CNN descrtiptor is only extracted from color images.

\subsection{Distance matrices with GNSS priors}
The extracted multiple descriptors $\{\textbf{d}^{f,m}|f\in{F},m\in{M}\}$ are leveraged to measure the similarity between images, thus to characterize the correspondence of query images and database images. In this paper, sequential images, rather than single images, are utilized during image matching. Assuming that the query sequence has the size of \textit{n} and the database sequence has the size of \textit{l}, then the distance matrix features the size of $ n \times l $. Herein, we define $ D^{f,m} $ as the distance matrix of descriptor \textit{f} extracted from modality \textit{m}. The element $ D^{f,m}_{i,j} $ of the matrix $ D^{f,m} $ is attained by measuring the descriptor distance between the \textit{i}-th query image and the \textit{j}-th database image. The distance measurement varies from different descriptors. For binary descriptors (LDB), Hamming distance is measured as the distance of images, while the distances of GIST and CNN descriptors are measured with Euclidean distance. 

Despite with the insufficient positioning accuracy, GNSS data consisting the coordinates of longitude and latitude provide with a priori knowledge for visual place recognition. With the GNSS priors, those query-database pairs that leave a large spatial distance between each other need not be matched, so as to improve the computational efficiency and to reduce the possibility of image mismatching. The metric distance between the \textit{i}-th query image and the \textit{j}-th database image is specified as $ G_{i,j} $, hence the final distance matrix containing GNSS data $ E^{f,m} $ is obtained by
\begin{equation}
E^{f,m}_{i,j} = \left\{
\begin{array}{ccc}
D^{f,m}_{i,j} &~~~~~& G_{i,j} \leq g,\\
Inf. &~~~~~& G_{i,j} > g.
\end{array} \right.,
\end{equation}
where $g$ is the threshold of possible matching pairs. The smaller the threshold $ g $, the smaller the searching range of image matching. Considering the observation error of the GNSS module used in this paper, the threshold $ g $ should not be too small, the correct matching results would be ruled out otherwise. In this paper, $ g $ is set to 15 meters. 

\subsection{Online sequence-base searching and matching scoring}
Having obtained a distance matrix $ E^{f,m} $, we execute an online cone-based searching upon every query-database pair, which achieves sequential image matching and gets a matching score for each pair. 

\begin{figure}
	\centering
	\includegraphics[width=0.9\columnwidth]{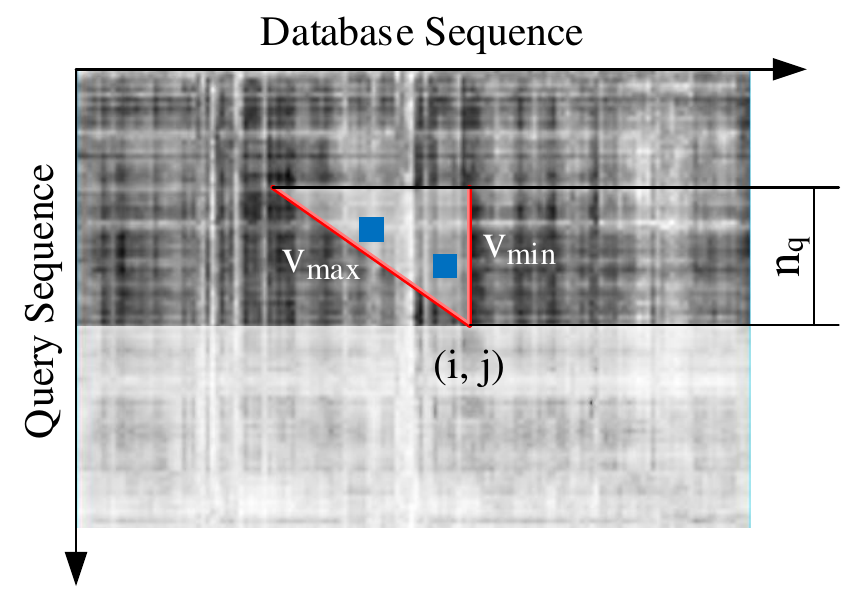}
	\caption{The online cone-based searching schematics.}
	\label{fig:3}
\end{figure}
As shown in Figure~\ref{fig:3}, the horizontal axis denotes the database sequence, and the vertical axis denotes the query sequence. Within the distance matrix, each query-database pair $(i,j)$ is associated with only one cone region which is limited by sequential length $ n_{q} $, maximal velocity $ v_{max} $ and minimal velocity $ v_{min} $. The online cone-based searching algorithm proposed in this paper is different from the offline one in~\cite{ConeSearch}. The offline searching algorithm makes use of the ``future'' query images, thus place recognition cannot run in real time.

Within the region, the number of best-matching pairs (represented by blue squares in Figure~\ref{fig:3}) is counted firstly. The best-matching pair is defined as the minimum value of a certain row in the distance matrix. In other words, a query descriptor and the database descriptor featuring minimum distance with that query descriptor compose a best-matching pair. Herein, the number of best-matching pairs in a cone region is defined as $ n_{match} $, and the score $ s_{i,j} $ of the query-database pair $(i,j)$ is defined as
\begin{equation}
s_{i,j} = \frac{n_{match} }{ n_{q}} .
\end{equation}

Naturally, all of the matching scores $ s_{i,j} $ form into a score matrix $ S^{f,m} $. Multiple descriptors extracted from different modalities carry diverse visual information, so assigning the same weight to different descriptors during image matching does not necessarily promote the matching robustness. Therefore, the coefficients of score matrix synthesis $\{\lambda^{f,m}\}$ need to be adjusted for the better accuracy of place recognition. The score matrices derived from different descriptors of different modalities are synthesized to a single score matrix $ S $, which is presented as
\begin{equation}
S_{i,j}=\frac{\sum\limits_{f\in{F},m\in{M}}\lambda^{f,m}\times S^{f,m}_{i,j}}{\sum\limits_{f\in{F},m\in{M}}\lambda^{f,m}}.
\end{equation}

The genetic algorithm~\cite{openga} is used to determine the values of $\{\lambda^{f,m}\}$, which is described later in Section~\ref{Experiments}. With matching score matrix, each query image corresponds to the best database image with the highest score. In order to get the final place recognition results, the matching score of the best query-database pair is evaluated to rule out the mismatching pairs. In this paper, we use score thresholding~\cite{OpenSeqSLAM2.0} to remove low-confidence matching results whose score lower than threshold $ t $.

\subsection{Implementation}
The proposed OpenMPR algorithm is implemented in C++, considering the portability and effectiveness. The open-source code of OpenMPR is available online~\cite{OpenMPR}. The dependencies include \textit{OpenCV 4.0}~\cite{OpenCV}, as well as \textit{DBoW3}~\cite{DBoW3} for BoW extraction, \textit{LibGIST}~\cite{LibGIST} for GIST extraction, \textit{libLDB}~\cite{LDB} for LDB extraction, and \textit{OpenGA}~\cite{openga} for parameter tuning.

The settings of OpenMPR could be easily switched by configuration file \textit{Config.yaml}. There are two modes in OpenMPR, which is testing mode and tuning mode. In testing mode, the place recognition is executed by using the default or customized parameters. In tuning mode, the optimal parameters are searched to achieve the best performance. The other configurable parts of OpenMPR involve the resolution of input images, whether to use GNSS data, and whether to use certain image modality or image descriptor.

\section{Experiments}
\label{Experiments}
In this section, the real-world place recognition dataset collected by the assistive device is illustrated firstly. In order to achieve the optimal performance of place recognition, the experiments on parameter tuning were carried out and the tuning results are analyzed thoroughly. Finally, the state-of-the-art performance of OpenMPR is validated through comparative study.

\begin{figure} 
	\begin{minipage}{\columnwidth}
		\centering
		\includegraphics[width=\textwidth]{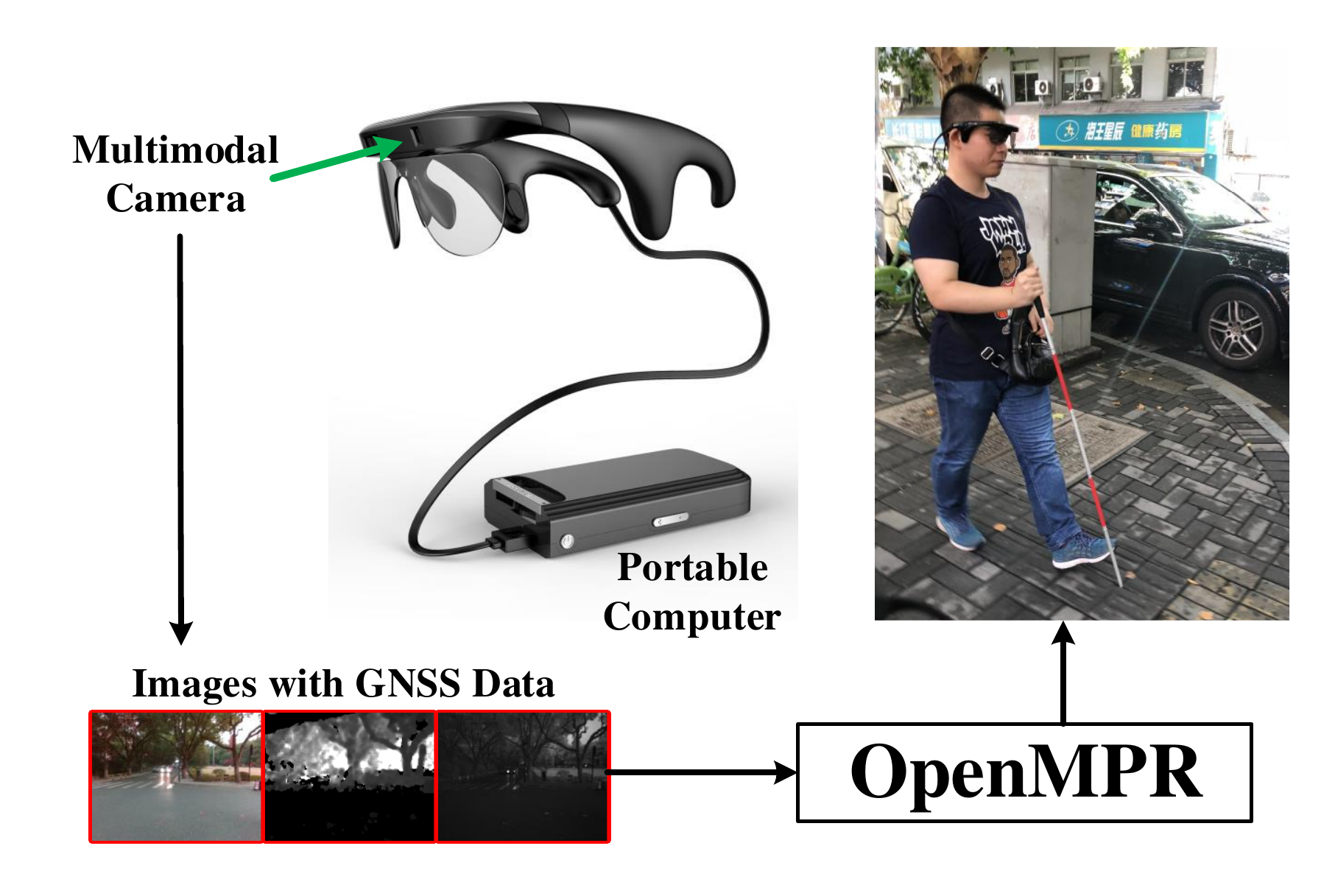}
	\end{minipage}
	\caption{The assistive device Intoer is used to capture multimodal data.}
	\label{fig:Intoer}
\end{figure}

In view that OpenMPR is prone to be implanted into assistive devices, the experiments were carried out on the assistive device Intoer~\cite{KrVision}, which is shown in Figure~\ref{fig:Intoer}. The assistive device Intoer is utilized not only to capture multimodal images and GNSS data but also to run the OpenMPR algorithm.

\subsection{Datasets}

In view that the place recognition dataset with multimodal data has not been released, we collected a real-world dataset, available at~\cite{OpenMPR}, within the campus of Yuquan, Zhejiang University. 

One frame of data consists of a color image, a depth image, an infrared image, and a GNSS coordinate. The multimodal images were collected using Intel RealSense ZR300 Camera~\cite{Realsense} embedded in Intoer, which is an infrared assisted stereo vision camera. In terms of the effective range and density of depth images, Realsense ZR300 represents the moderate level among commercial RGB-D cameras. Thereby, the dataset proposed in this paper is adequate to evaluate the performance of OpenMPR. Other imaging specifications are illustrated in Table~\ref{images}. The GNSS data were collected with the customized GNSS receiver embedded in Intorer.

\begin{table}
	\centering
	\caption{The specifications of multimodal images captured by ZR300.}
	\label{images}
	\begin{tabular*}{\columnwidth}{@{\extracolsep{\fill}}cccc@{}}
		\hline
		  & Color & Depth & Infrared\\
		\hline	
		Resolution & $320\times240$       & $320\times240$ & $320\times240$  \\	
		Shutter type          & Rolling       & Global   &  Global     \\
		Frame rate & 1 FPS       & 1 FPS & 1 FPS\\		 	
		\hline
	\end{tabular*}
\end{table}

\begin{figure} 
	\begin{minipage}{\columnwidth}
		\centering
		\includegraphics[width=\textwidth]{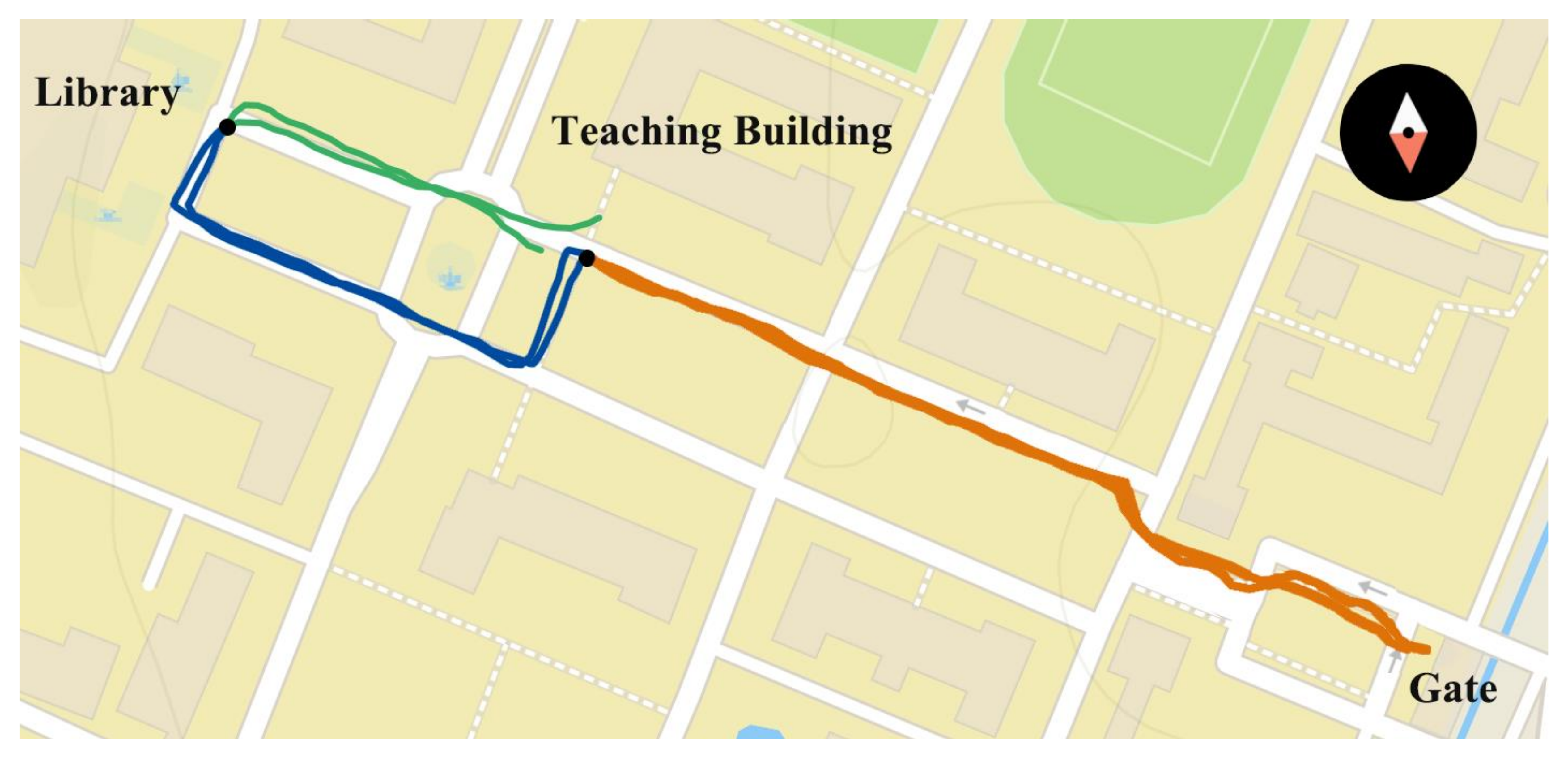}
	\end{minipage}
	\caption{The three traversed paths in multimodal datasets of place recognition. Route A: from the teaching building to the gate (orange). Route B: from the teaching building to the library (green). Route C: from the library to the teaching building (blue).}
	\label{fig:path}
\end{figure}

Up to 1,671 frames of data are involved in the dataset, where the four subsets were collected on three routes as shown in Table~\ref{dataset} and Figure~\ref{fig:path}. Although Train-1 and Train-2 cover the same route, they were collected in the opposite traversing direction. It is worthwhile to note that no route overlap exists among the training subsets and the testing subsets. Moreover, images of the four subsets are not selected artificially. In the experiments, Train-1 and Train-2 are utilized to tune the parameters, and Test-3 and Test-4 are used to validate the performance of multimodal place recognition.  

Each subset is composed of one query sequence and one database sequence. The collected multimodal images feature apparent viewpoint changes between query and database sequence, since the camera is embedded in the wearable device and the query and database were not captured on the completely identical route. Apart from that, all of the images also present dynamic object changes between query and database. For example, the person passing by in front of the camera appears in the query sequence, but does not appear in the database sequence. Moreover, the illumination changes exist in Train-2 and Test-4. In those subsets, database and query images were captured in the afternoon and at dusk separately. All of those real-world changes form into substantial challenges for place recognition. In brief, the dataset was collected in real-world scenarios, and is suitable for the localization issues of assistive technology.  

Due to the inaccuracy of GNSS data, the ground truth of dataset is labeled manually according to visual similarity rather than GNSS distance. Each query image is associated with the best-matching database image.  
\begin{table}
	\centering
	\caption{The characteristics of multimodal place recognition dataset. v = viewpoint changes, o = dynamic objects, and i = illumination changes. }
	\label{dataset}
	\begin{tabular*}{\columnwidth}{@{\extracolsep{\fill}}ccccc@{}}
		\hline
		Subset&Route  & \# query & \# database&Changes\\
		\hline	
		Train-1 &C& 212       & 291&v/o \\
		Train-2	&C&208&215&v/o/i\\
		\hline
		Test-3  &A         & 97       & 142   &  v/o     \\
		Test-4 &A+B& 233       & 273 & v/o/i\\		 	
		\hline
	\end{tabular*}
\end{table}

\subsection{Parameter tuning procedures and results}
In order to optimize the performance of place recognition, a series of parameters are tuned on the training datasets that are separated from the testing datasets. The parameters include the length ($ n_{q} $) and velocity limits ($ v_{max} $ and $ v_{min} $) of cone searching, the coefficients of score matrix synthesis ($ \lambda^{f,m} $), and the threshold $ t $ of score thresholding. 

If a query image matches with a database image, the result is defined as a positive result. If no database image is matched, the result is defined as a negative result. Considering the query and database images are sequential in the dataset, the place recognition result of a query image could be represented as the sequential index of the best-matching database image. If the index difference between the place recognition result and the ground truth is less than or equal to the tolerance (set to 5 in this paper), the result is defined as a TP (true positive) result. Otherwise, the positive result is defined as a FP (false positive) result. Moreover, if the result should match with a database image but it does not match with any database image, the result is defined as a FN (false negative) result. The performance of OpenMPR is evaluated and analyzed in terms of precision and recall. Precision is the proportion of true positives out of all predicted positives, and recall is the proportion of true positives to all of actual positives. 
\begin{equation}
Precision=\frac{TP}{TP+FP}
\end{equation}
\begin{equation}
Recall=\frac{TP}{TP+FN}
\end{equation}
In this section, the objective of parameter tuning is to choose the parameters that achieve the greatest $ F_1 $ score. 
\begin{equation}
F_1=2\times\frac{Precision\times{Recall}}{Precision+Recall}.
\end{equation}

The following section outlines the procedures and results of configurable parameter tuning. 
\subsubsection{Coefficients of score matrix synthesis}
\label{coeffTuning}
The coefficient $ \lambda^{f,m} $ denotes the importance of specific descriptor $ \textbf{d}^{f,m} $ during place recognition. In order to tune the coefficients efficiently, we leverage the genetic algorithm implemented by~\cite{openga} to seek the optimal combination of coefficients. The length of the cone region ($ n_{q} $) is set to 1, in order that the sequential searching does not affect coefficient optimization.

The genetic algorithm is an analogue of natural selection, which optimizes the parameters (genes) by the bio-inspired operators such as mutation, crossover and selection. The coefficient array with the size of 9 (see Table~\ref{coeff}) is defined as the genes, and the fitness of genes is evaluated with the $ F_1 $ score at that coefficient combination. The principles and implementation details of the genetic algorithm could be found in~\cite{openga}. Empirically, the maximum number of generations (80 in this paper) is set as the stopping criterion, which is sufficient for the genetic algorithm to generate a stable iterative result. The genetic algorithm runs for multiple times (15 in this paper) with randomly initialized genes to avoid the local optima. The mean coefficients of the multiple results obtained by the genetic algorithm are set as final parameter searching results. The mean coefficients of the two datasets are chosen as the optimal parameters, as presented in Table~\ref{coeff}.

\begin{table*}
	\caption{The searching results of coefficients $ \lambda^{f,m} $ using the genetic algorithm. (c=color, d=depth, i=infrared.)}
	\label{coeff}
	\begin{tabular*}{\textwidth}{@{\extracolsep{\fill}}cccccccccccc@{}}
		\hline
		\multirow{2}{*}{Dataset} & \multicolumn{2}{c}{$f=BoW$}&\multicolumn{3}{c}{$f=GIST$}&\multicolumn{3}{c}{$f=LDB$}&	\multirow{2}{*}{$\lambda^{CNN,c} $}&\multirow{2}{*}{$ F_1 $}\\
		& $ \lambda^{f,c} $  & $\lambda^{f,i} $ & $\lambda^{f,c} $ & $\lambda^{f,d} $  &  $\lambda^{f,i} $&
		$\lambda^{f,c} $ & $\lambda^{f,d} $ &  $\lambda^{f,i} $&\\
		\hline
		Train-1 &1.359&	1.666&	2.269&	1.102&	0.986&	0.617&	0.469&	1.042&	0.491&  0.73	\\
		Train-2 &1.132&	1.705&	0.889&	1.081&	0.989&	0.436&	0.778&	0.638&2.353		& 0.63\\
		\hline
		Optimal &1.245&	1.685&	1.579&	1.091&	0.987&	0.526&	0.623&	0.840&	1.422	&-
		\\
		\hline
	\end{tabular*}
\end{table*}

As demonstrated in Table~\ref{dataset}, Train-1 suffers from viewpoint changes and dynamic objects, meanwhile Train-2 suffers from more illumination changes than Train-1. On Train-2, the CNN descriptor presents the highest weights compared with other descriptors, which illustrates that the descriptors derived from the GoogLeNet pre-trained on Places365 yield superior description performance even under the severe changes. On dataset 1, the GIST descriptors show better description performance compared with other descriptors, which indicates that GIST descriptor is suitable for depicting the images without large illumination changes. Besides, LDB presents the suboptimal performance of place recognition in the complicated environments. The dataset used in this paper features severe viewpoint changes and dynamic objects, hence the holistic descriptors are important to grasp the global information.

Compared with the holistic descriptors, the performance of BoW descriptors on the two datasets reveals that BoW is advantageous and stable for place recognition in various environments. More conclusions can be drawn when inspecting the results on color and infrared modality carefully. The BoW descriptor on the infrared modality features a higher weight than that on color modality. The reasonable explanation is that the local ORB features are susceptible to image details with motion blur, which is prone to occur on color images captured with the rolling shutter. On the contrary, the infrared modality features better performance on imaging stability thanks to the global shutter.



\subsubsection{Parameters of cone-based searching}
Having chosen the optimal coefficients (shown in Table~\ref{coeff}), the tuning procedures of other parameters are executed. The parameters of cone-based searching algorithm include the length of sequence ($ n_{q} $) and velocity limits ($ v_{max} $ and $ v_{min} $). They represent the quantity of information used in cone searching. In parameter sweeping, the maximal velocity ($ v_{max} $) is set as the reciprocal of the minimal velocity ($ v_{min} $), so there are only two parameters to be tuned. The minimal velocity ($ v_{min} $) is varied from 0.1 to 0.75, and the length ($ n_{q} $) is varied from 3 to the 79. 

\begin{figure} 
	\begin{minipage}{\columnwidth}
		\centering
		\includegraphics[width=\textwidth]{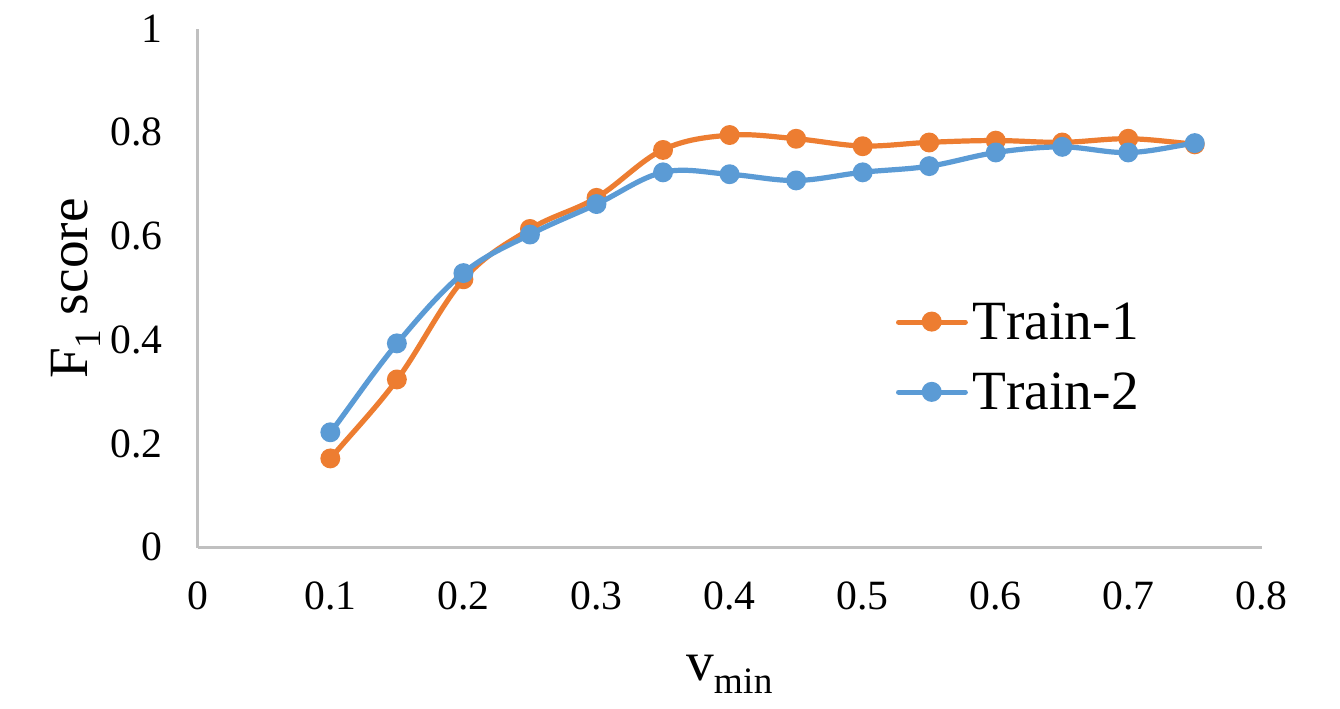}
	\end{minipage}
	\caption{The parameter sweeping results of $ v_{min} $.}
	\label{fig:v}
\end{figure}

Different velocity limits of cone-based searching are utilized to test the performance of place recognition. The Figure~\ref{fig:v} demonstrates that $ v_{min}\ge0.4 $ ($ v_{max}\le2.5 $) features good performance, and that the larger velocity range results in suboptimal performance. The large searching range introduces more best-matching pairs, meanwhile introduces more potential inaccurate results. The velocity limits should be moderate to tolerant the real-world scenarios, such as the inconsistency of the carrier's walking speed when recording query and database sequences. Thereby, we set  $ v_{min}=0.4 $, and $ v_{max}=2.5 $.

\begin{figure}
	\begin{minipage}{\columnwidth}
		\centering
		\includegraphics[width=\textwidth]{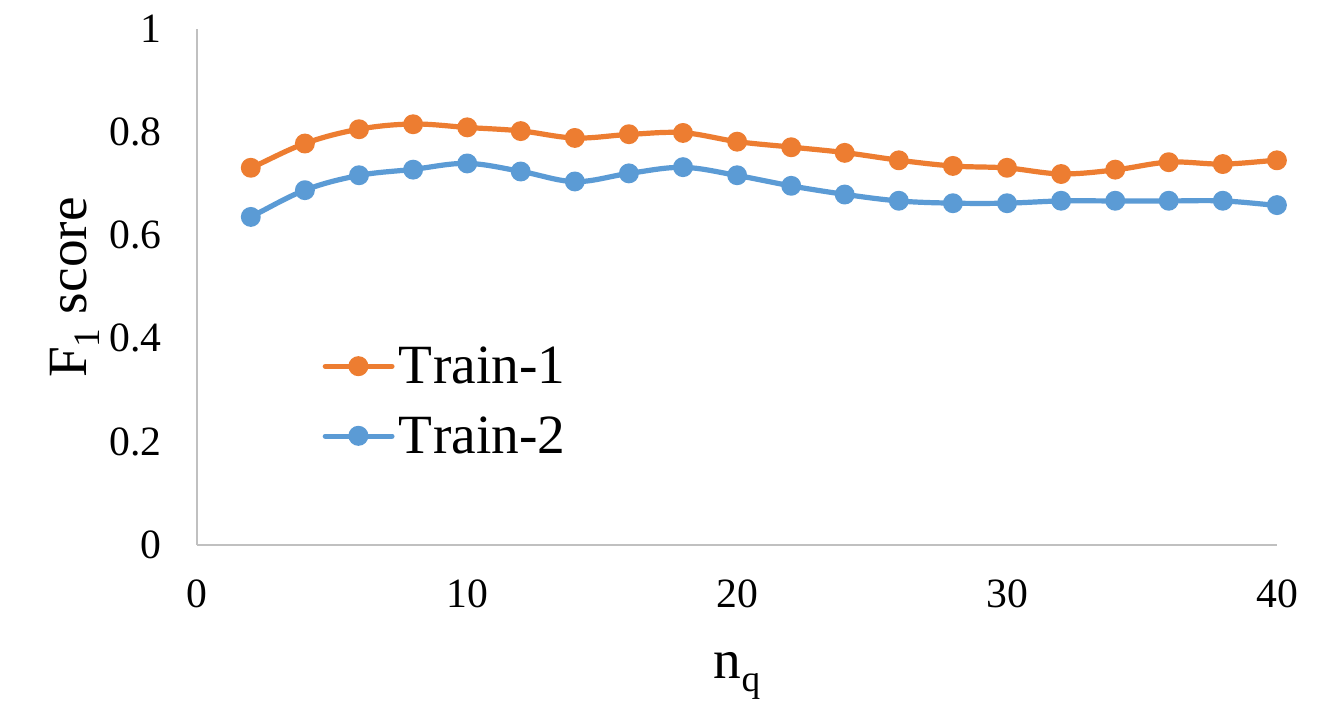}
	\end{minipage}
	\caption{The parameter sweeping results of $ n_{q} $.}
	\label{fig:n}
\end{figure}

As shown in Figure~\ref{fig:n}, the performance of place recognition is related to the length of searching sequence ($ n_{q} $). Whether $ n_{q} $ is too large or too small, the performance is limited. For the sake of computational efficiency, we set the optimal parameter $ n_{q} $ as 10.

\subsubsection{Threshold of score thresholding}
\begin{figure}
	\begin{minipage}{\columnwidth}
		\centering
		\includegraphics[width=0.9\textwidth]{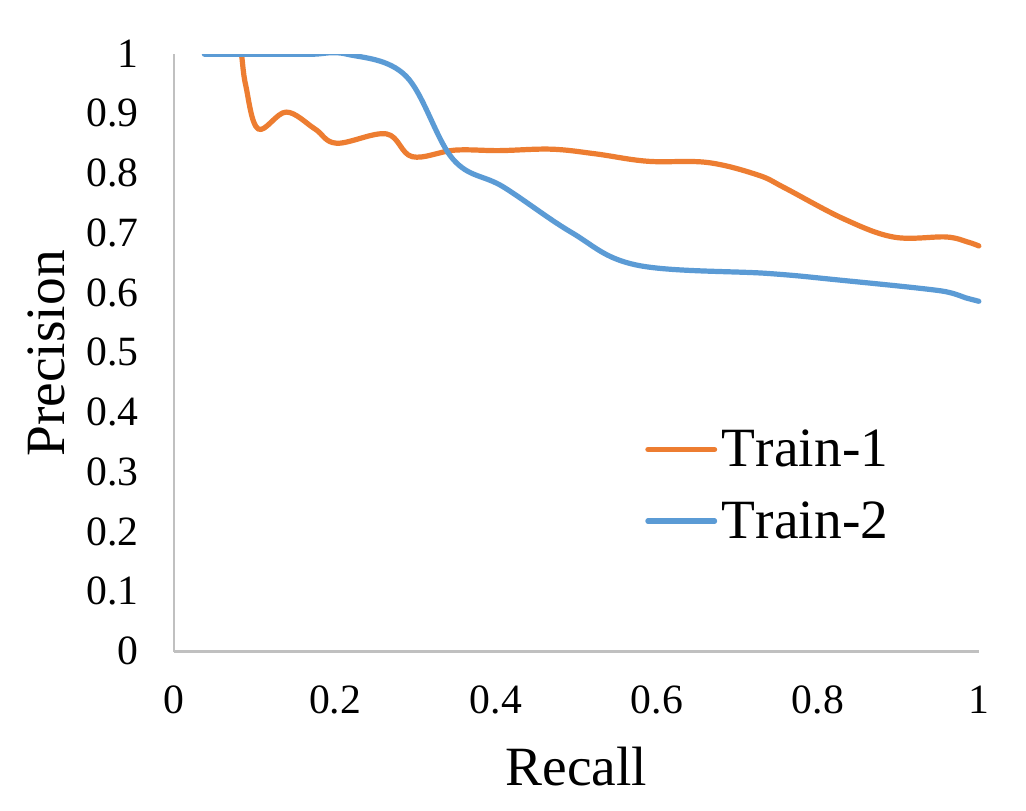}
	\end{minipage}
	\caption{The precision-recall curve as sweeping parameter $ t $.}
	\label{fig:t}
\end{figure}

The threshold score thresholding $ t $ affects the precision and recall of place recognition.  The score threshold $ t $ is used to eliminate bad matching results and improve the performance of place recognition. As shown in Figure~\ref{fig:t}, the precision-recall curve under different thresholds $ t $ is plotted. As the threshold is low, the matching results with low confidence influence the precision of place recognition. On the contrary, the high threshold results in low recall rate. The optimal value of threshold $ t $ is set to 0.16, where the recall has not descended substantially and the precision maintains at a high level. 

\subsection{Validation of OpenMPR}
In order to validate the parameter tuning results and the systematic performance of OpenMPR, the testing sets whose routes are different from those of training sets are utilized to evaluate the proposed algorithm. As demonstrated in Table~\ref{dataset}, viewpoint changes and dynamic objects exist in both subsets, meanwhile illumination changes are introduced in Test-4.

\subsubsection{Validation of optimized coefficients}
To validate the effectiveness of optimized coefficients $\{\lambda^{f,m}\}  $, the place recognition performance on different coefficient configurations is compared. In addition to optimized coefficients, the other configuration involves

 (1) $ \lambda=1 $: let all of the coefficients be 1, which means that all descriptors feature the same importance. 
 
 (2) \textit{w/o certain descriptor}: let the coefficients of the corresponding descriptor be 0, which means abandoning that descriptor in the system. 
 
\begin{figure}
	\begin{minipage}{\columnwidth}
		\centering
		\includegraphics[width=\textwidth]{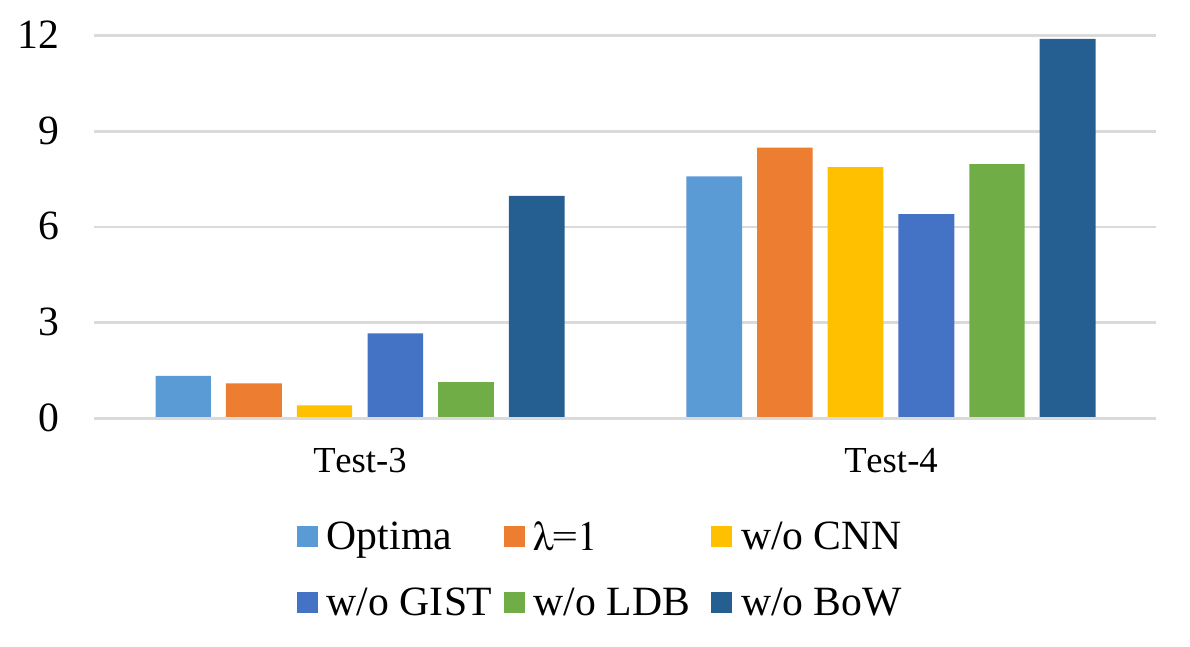}
	\end{minipage}
	\caption{The mean localization error under different configurations of parameters.}
	\label{fig:config}
\end{figure}

 The mean localization error of different coefficient configurations is shown in Figure~\ref{fig:config}. Herein, the localization error refers to the index difference between the OpenMPR result and the ground truth. The mean localization error is obtained by averaging the localization error of all query images in the testing set. It is concluded that the configuration of optimized coefficients shows the balanced performance on both testing sets, though it is not the best configuration on the single dataset. 

On both testing sets, the configuration \textit{w/o BoW} features the worst performance, which illustrates that BoW descriptor is essential for place recognition. On Test-3, the configuration \textit{w/o CNN} yields the optimal performance, and the configuration \textit{w/o GIST} shows the suboptimal performance. Those phenomena are consistent with the analysis in Section~\ref{coeffTuning} that the GIST descriptor, instead of the CNN descriptor, plays the vital role in place recognition if there is no illumination changes. In contrast, on the testing set with illumination changes, the GIST descriptor is no longer eligible for good performance, the CNN descriptor and the other descriptors are indispensable.

\subsubsection{Validation of systematic performance}
With the optimal parameters determined in the preceding sections, the place recognition results of OpenMPR on the two testing sets are compared with the state-of-the-art place recognition algorithms. OpenSeqSLAM2.0~\cite{OpenSeqSLAM2.0} and Visual Localizer~\cite{VisualLocalizer} are chosen as the baselines of OpenMPR. Though OpenSeqSLAM2.0 was designed for visual place recognition on autonomous vehicles, it provides with important inspirations in terms of sequence searching and matching selection techniques. In the experiments, the OpenSeqSLAM2.0 parameters related to sequence searching and matching selection were set to those optimal values presented above. As a preliminary work, Visual Localizer proposed a place recognition solution for the mobility of visually impaired people using pretrained CNN descriptors and global optimization. 

\begin{table}
	\centering
	\caption{The place recognition results on the testing datasets.}
	\label{test}
	\begin{tabular*}{\columnwidth}{@{\extracolsep{\fill}}ccccc@{}}
		\hline
		Algorithm&	Subset  & Precision & Recall & Error\\
		\hline	
		\multirow{2}{*}{OpenMPR}&	Test-3 & 88.7\%  & 100.0\%&\textbf{1.33} \\	
		&	Test-4	&57.8\% &99.3\%&\textbf{7.59}\\
		\hline
		OpenSeq-&	Test-3 & 26.6\%  & 34.0\%&7.80 \\	
		SLAM2.0&	Test-4	&25.7\% &82.0\%&47.91\\
		\hline		
		Visual&	Test-3 & 48.5\%  & 100\%&19.14 \\	
		Localizer&	Test-4	&58.4\% &100\%&9.99\\
		\hline
	\end{tabular*}
\end{table}

As shown in Table~\ref{test}, the three performance indicators (precision, recall and mean localization error) are leveraged to evaluate the results of OpenMPR on the two testing sets. In terms of mean localization error, the proposed OpenMPR is superior to the two state-of-the-art algorithms. According to the statistics of OpenMPR, the place recognition on Test-3 is more precise than that on Test-4, in that fewer appearance changes are involved in Test-3. Fortunately, with the help of the multiple descriptors extracted from multimodal images, the mean localization error of OpenMPR on Test-4 is acceptable, which slightly exceeds the tolerance of 5. GNSS priors play an important role in ruling out the definite negatives during image matching.

Compared with OpenMPR, OpenSeqSLAM2.0 yields inferior localization performance both on Test-3 and Test-4, in view of the low recall and large localization error. Apparently, OpenSeqSLAM2.0 that measures the similarity of images via sum of absolute differences of normalized images does not make place recognition robust against various appearance changes. The comparison between OpenMPR and OpenSeqSLAM justifies that the proposed image descriptors robustify place recognition under practical conditions. For Visual Localizer, the performance of Test-4 with more appearance changes surpasses that of Test-3, which resembles the phenomenon that CNN descriptor features a higher weight on Train-2 than on Train-1. It is evident that the proposed CNN descriptor (the compressed concatenation of $inception3a/3\times3$ and $inception3a/3\times3\_reduce$) is capable of extracting effective semantic ``place fingerprint'' between images with large appearance changes. However, without the aid of other descriptors and multimodal images, the performance of the CNN descriptor on Test-3 is limited, which further confirms that the necessity of multiple descriptors proposed in this paper.    

\begin{figure*}
	\centering
	\includegraphics[width=\textwidth]{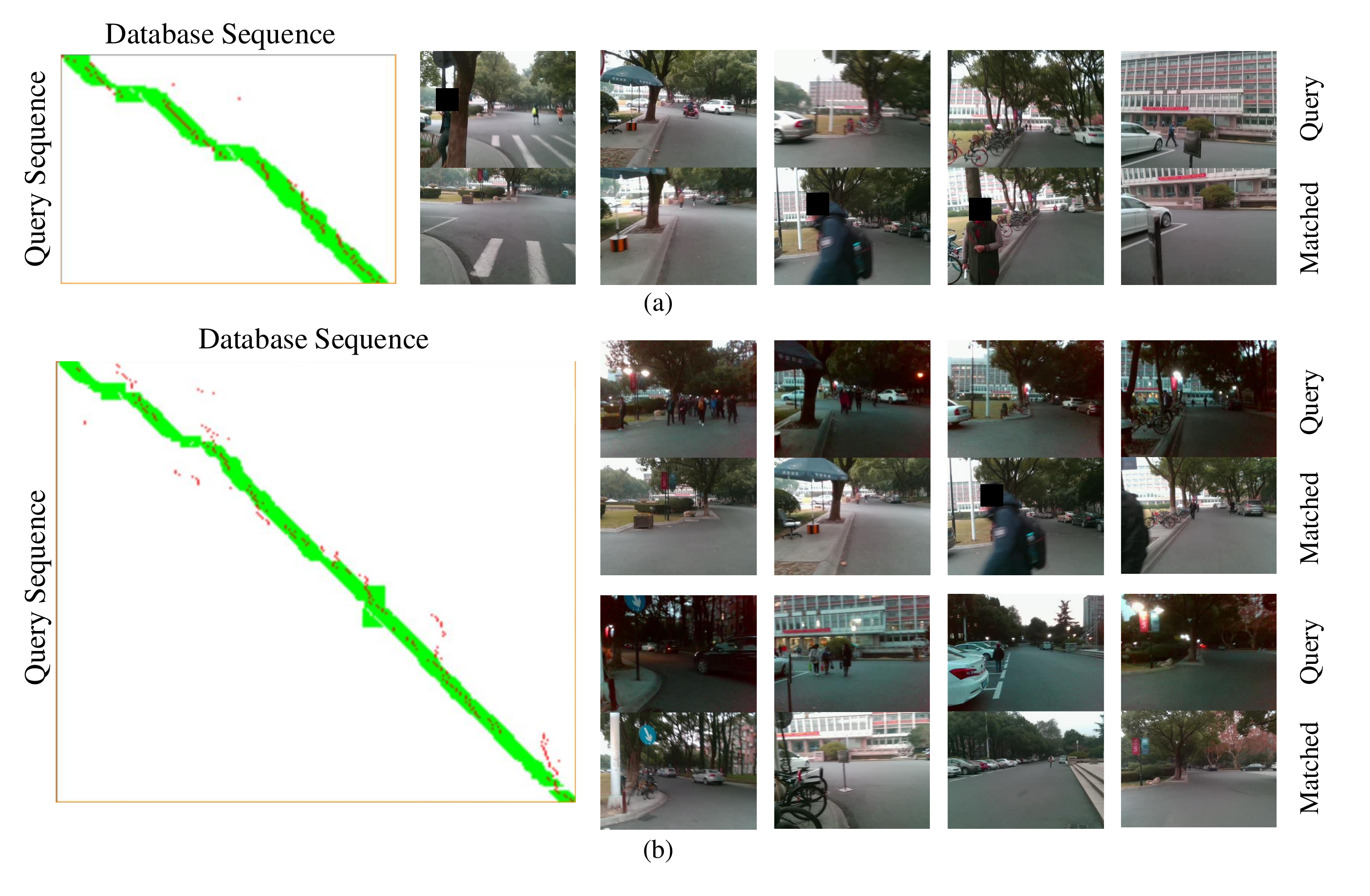}
	\caption{The place recognition results and some localization instances on (a) Test-3 and (b) Test-4 dataset. In the left diagram, the horizontal axis denotes the database sequence, and the vertical axis denotes the query sequence.}
	\label{fig:res}
\end{figure*}
As shown in Figure~\ref{fig:res}, the place recognition result of OpenMPR is visualized as a visualization matrix with the size of $n\times{l}$, where $n$ is the number of query images and $l$ is the number of database images. The element of the matrix denotes the query-database pair. In the matrix, green and red points denote ground truths (with the tolerance of 5) and localization results respectively. From the diagrams, it is concluded that the place recognition results basically conform to the corresponding ground truths, despite the serious viewpoint variation, motion blur and dynamic objects (e.g. pedestrians). Even on Test-4 with obvious illumination changes, most of the mismatching images are not far from the tolerance of place recognition. In Figure~\ref{fig:res}, some successful matching results are presented, which indicates that OpenMPR still recognizes places under the conditions of various appearance changes.

Real-time performance is crucial for assistive navigation. OpenSeqSLAM2.0 uses both the ``past'' images and the ``future'' images during cone-based searching, hence it cannot be used in real time. Unfortunately, network flow-based global optimization scheme embedded in Visual Localizer features inferior computational efficiency. The single-frame computation speed is analyzed on the Inoter with Intel Atom x5-Z8500 and a desktop with Inter Core i5-6500 to evaluate the real-time performance of OpenMPR, as shown in Table~\ref{Timing}. The real-time requirement is basically satisfied by OpenMPR according to the results on the Intoer. With the update of Intoer hardware, the real-time performance would be further improved in view of the speed test on the desktop. After inspecting the running time of descriptor extraction, it is found that time consumed during extracting GIST descriptors from multimodal images accounts for the major proportion (more than 80\% of descriptor extraction). In the future, applying GIST library with superior computational efficiency to OpenMPR leads to better real-time performance of the system.

\begin{table}
	\centering
	\caption{The real-time performance of OpenMPR on different platforms.}
	\label{Timing}
	\begin{tabular*}{\columnwidth}{@{\extracolsep{\fill}}cccc@{}}
		\hline
\multirow{2}{*}	{Platform}&	Descriptor    & \multirow{2}{*}{Matching} & \multirow{2}{*}{Overall}\\
		&Extraction&&\\
		\hline	
		Intel Atom &\multirow{2}{*}{2,056 ms} & \multirow{2}{*}{98 ms}  & \multirow{2}{*}{2,154 ms}\\	
		x5-Z8500&	& & \\
		\hline		
		Intel Core&	\multirow{2}{*}{362 ms} & \multirow{2}{*}{25 ms} & \multirow{2}{*}{387 ms} \\	
		i5-6500& 	&  & \\
		\hline
	\end{tabular*}
\end{table}

\section{Conclusion}
Different with the majority of place recognition work, this paper focus on the traveling demands of visually impaired people, and propose an open-source software OpenMPR, which leverages multi-modal data for online place recognition task.

In the area of assistive technology, the wearable camera tends to capture images with motion blur and low resolution. Due to the limited computational resource, discrete images (one image per second in this paper), instead of video streams, are captured and processed on the portable devices. Apart from that, the query and database sequences features various appearance changes, including viewpoint changes, illumination changes and dynamic objects. In those real-world scenarios, the proposed OpenMPR utilizes configured multiple descriptors extracted from multimodal data and online sequence-based searching to obtain good place recognition performance. It achieves 88.7\% precision at 100\% recall without illumination changes, and achieves 57.8\% precision at 99.3\% recall with illumination changes.

In the future, we plan to achieve semantic place recognition, where the visual information in images is understood and places are autonomously labeled with different levels of importance.

\section{Acknowledgments}
This work was supported by the State Key Laboratory of Modern Optical Instrumentation.

\section{References}
\bibliographystyle{dcu}
\bibliography{ref} 

\end{document}